\documentclass[runningheads]{llncs}

\usepackage{eccv}

\usepackage{eccvabbrv}

\usepackage{graphicx}
\usepackage{booktabs}
\usepackage{amsmath}
\usepackage{amssymb}
\usepackage{siunitx}
\usepackage{xcolor}
\usepackage{subcaption}
\usepackage{multirow}
\usepackage{pifont}    
\usepackage{placeins}  

\usepackage[pagebackref,breaklinks,colorlinks,citecolor=eccvblue]{hyperref}

\DeclareSIUnit{\px}{px}
\newcommand{\ind}{\mathbf{1}}

\definecolor{maleblue}{rgb}{0.0,0.59,0.90}
\definecolor{femgreen}{rgb}{0.31,0.78,0.31}
\definecolor{juvorange}{rgb}{1.0,0.55,0.0}
\newcommand{\male}{\textcolor{maleblue}{\textbf{male}}}
\newcommand{\female}{\textcolor{femgreen}{\textbf{female}}}
\newcommand{\juv}{\textcolor{juvorange}{\textbf{juvenile}}}

\begin{document}

\title{When One Modality Is Not Enough:\\ Multimodal Sex and Life-Stage
Classification of Red Deer from Aerial RGB--Thermal Video}
\titlerunning{When One Modality Is Not Enough}

\newcommand{\equalcontrib}{\textsuperscript{*}}

\author{
  Hugo Markoff\inst{1}\equalcontrib \and
  Christoph Praschl\inst{2}\equalcontrib \and
  Ivan Ludo\v{s}ki\inst{3} \and
  Sara Beery\inst{4} \and
  Michael {\O}rsted\inst{1} \and
  David C. Schedl\inst{2}
}

\authorrunning{H. Markoff et al.}

\institute{
  Aalborg University, Department of Chemistry and Bioscience, Aalborg, Denmark
  \and
  University of Applied Sciences Upper Austria, Hagenberg, Austria
  \and
  University of Novi Sad, Faculty of Agriculture, Novi Sad, Serbia
  \and
  Massachusetts Institute of Technology, Cambridge, MA, USA
}

\maketitle

\begingroup
\renewcommand{\thefootnote}{*}
\footnotetext{Corresponding authors (equal contribution): Hugo Markoff (khbm@bio.aau.dk), Christoph Praschl (christoph.praschl@fh-hagenberg.at)}
\endgroup

\begin{abstract}
Aerial drone surveys increasingly support wildlife population estimation, yet a useful census is more than a count: population dynamics are defined by species composition, sex ratios and age structure, that is, by which species are present and how a herd splits into adult males, adult females and juveniles. We use red deer (\textit{Cervus elaphus}) as a test case, because managers act on these dynamics and because the visible cue defining adult males, the antlers, is seasonally variable. Surveys are flown nadir, the cameras looking straight down while the drone flies high enough not to disturb the animals, so each deer occupies only a small, low-resolution patch. The two recording modalities fail in opposite conditions: in color a deer under canopy blends into the ground, while in thermal it becomes a bright blob that loses fine detail. Rather than trust either modality alone, we fuse them at every stage using self-supervised DINOv3 features. Our pipeline tracks animals in both modalities, treats an animal as confirmed only when the two cameras agree, keeps only the clear, non-occluded frames, and assigns species and sex by a vote across them; life stage is read separately from geo-referenced body size, since at survey resolution a juvenile often only differs from an adult female in size. Across four flights spanning the antler season the fused pipeline correctly classifies 25 of the 26 detected individuals (7 of 8 adult males, all 16 adult females and 2 juveniles), against 20 of 26 for either sensor alone. The multimodal species classification reaches \SI{96.0}{\percent}, while for the sex classification, the fusion of the two sensors matters most: the combined RGB+thermal model is the most robust across environments and seasons. Automating the demographic classification turns a drone flight from a count into a repeatable reading of herd structure, so the sex ratios and age structure that managers already act on can be gathered as often as a survey can be flown.

\end{abstract}

\section{Introduction}
\label{sec:intro}

\begin{figure}[tb]
  \centering
  \includegraphics[width=\linewidth]{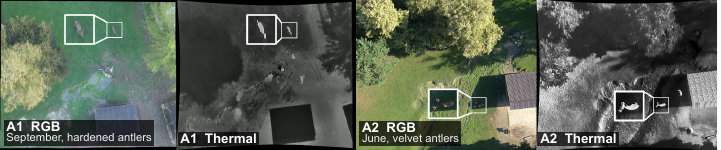}
  \caption{\textbf{Why one modality is not enough, and why it is seasonal.} One test area, two flights, each an RGB\,$|$\,thermal pair with the male enlarged. In \textbf{A1 (September)} the antlers have hardened and gone thermally dead, so the sex cue survives only faintly, and only in RGB. In \textbf{A2 (June)} deer resting in the shaded strip beside the barn are barely separable from the ground in RGB but bright in thermal, where the warm velvet antlers resolve.}
  \label{fig:teaser}
\end{figure}

Wildlife management needs to know not merely how many animals are present but \emph{which} animals: how a population divides into adult males, adult females and juveniles. A survey reads that composition as a snapshot, the age and sex structure from which recruitment, survival and population trend are afterwards estimated~\cite{clutton1982red}, yet it is exactly what detection and counting models do not report. We study the problem on red deer (\emph{Cervus elaphus}), where the male/female/juvenile split is a standard management variable and the cues are unusually well defined from above: adult males carry antlers and adult females do not, while juveniles can look like small adult females and are difficult to tell apart by appearance at survey resolution. Uncrewed aerial vehicles (UAVs) make repeated, low-disturbance surveys practical~\cite{gonzalez2016unmanned,corcoran2021automated}, and pairing an RGB with a thermal camera is attractive in forest habitats, where animals slip in and out of canopy and its shadow. The two modalities fail in \emph{opposite} conditions: in RGB a deer in cast shadow or under a crown is almost indistinguishable from the background, while in thermal it is an unmistakable bright blob but fur color is lost and antlers are visible only seasonally, limited by low resolution and warm-ground clutter. \Cref{fig:teaser} shows both failures, and which sensor performs better depends on the time of year: in June the velvet antlers are still warm and resolve in thermal, while by September they have hardened and gone thermally dark, so the cue survives only in RGB.

\paragraph{From rejecting cases to selecting frames.}
Prior work on the same data treated the demographic label as a question of ecological evidence as much as of modeling: on small, occluded nadir crops the sex cue is often simply absent from a given view, so the response is to abstain rather than guess~\cite{softprior2026}. It also showed that heads built on pretrained foundation-model representations beat smaller specialized architectures such as YOLO, whose weaker capacity struggles with the faint cues of aerial imagery. Abstaining matters because forcing ambiguous crops into a class injects label noise that corrupts the classifier~\cite{northcutt2021confident,geifman2017selective}, and clean supervision here is usually bought through active learning and review~\cite{kellenberger2019half,kellenberger2020aide}. But abstention has a price: rejected crops are often ambiguous animals, and a census cannot discard its hard cases. In video an animal is not one crop but a \emph{track} of many frames across the two modalities, so we reject the \emph{frames} on which the evidence is absent or limited and settle sex and life stage by voting over the survivors, which keeps the clean-label benefit without losing the individuals abstention would discard.

No single modality and no single frame is therefore enough for an individual-level census, yet almost all aerial wildlife pipelines stop at detection, counting or coarse tracking and leave the demographic step to manual review. We close that gap for one species as a demonstration that the step can be automated at all. The work extends a thermal-only localization and tracking pipeline~\cite{praschl2026stay} from one thermal stream to paired RGB--thermal detection and cross-modal tracking, from the native camera perspective to terrain-orthorectified imagery, and from localization to demographic classification. On the public BAMBI dataset (\cref{sec:data}) we contribute a pipeline that fuses the sensors at every stage and outputs, per individual, a species, a sex call, a life-stage flag and an occlusion-aware confidence, rendered onto the tracked video.

\paragraph{Scope and questions.} Four questions run through what follows. (i) What does cross-modal confirmation buy when deciding which tracks are animals and which are sensor noise (\cref{sec:track,sec:detres})? (ii) Which modality carries the sex signal (\cref{sec:cls,sec:sexres})? (iii) Does selecting frames rather than classifying all of them sharpen the census (\cref{sec:selective,sec:speciesres})? (iv) How reliable is a box-derived size cue as a life-stage biometric, and what limits it (\cref{sec:size,sec:discussion})?

\FloatBarrier
\section{Related Work}
\label{sec:related}

\subsection{Aerial wildlife datasets and fine-grained interpretation}
Two things separate our setting from most airborne wildlife work: the habitat is forested rather than open, and the output is a demographic profile rather than a count. Airborne monitoring has mostly been developed for open habitats, where animals are comparatively visible and RGB alone often suffices for localization, counting and tracking~\cite{gonzalez2016unmanned,corcoran2021automated,naik2024bucktales}. Recent methods cut annotation effort further by learning density-based localization from points rather than boxes~\cite{chacon2026overhead}, but do not recover the extent, orientation or appearance that individual tracking and demographic analysis need. Under canopy, RGB--thermal fusion improves robustness~\cite{bondi2020birdsai,krishnan2023fusion}, and thermal UAV surveys support red deer counting~\cite{zabel2023assessment} and antler-based sex-ratio estimation~\cite{ito2022antler}. All of these return aggregate quantities: a count for a flight, or a single ratio for a herd. None attaches a species, a sex and a life stage to an individual animal, and it is that step, from a population-level summary to a per-animal record, that we take here.

\subsection{Tracking and re-identification}
Online trackers of the SORT family~\cite{sort} pair a constant-velocity Kalman filter~\cite{kalman} with greedy association and remain a strong, simple baseline for aerial video. Appearance re-identification increasingly rests on self-supervised features: DINOv2 and DINOv3~\cite{oquab2023dinov2,dinov3} give transferable descriptors without labels, which triplet metric learning~\cite{triplet,hermans2017defense} can sharpen for identity and attribute discrimination. In ecology such metric learning underpins animal \sloppy{photo-identification~\cite{miele2021revisiting}}, though careful evaluation is needed to separate real gains from tuning artifacts~\cite{musgrave2020reality}.

Tracking alone does not scale to a survey: stitchers assume a static landscape, so animals that move during a flight are duplicated or ghosted in the orthomosaic and bias the count, which is why removing duplicates by distinguishing individuals is named an open direction~\cite{brack2025counting}, while thermal UAV work on ungulates stops at detection and counting~\cite{povlsen2024yolo,zabel2023assessment}. Aerial re-identification is usually approached from a lateral view, at least for species whose markings sit on the flank: Sun \etal~\cite{sun2026autonomous} pair YOLOv11 with a DINOv2 pose classifier but maneuver the drone around each animal to expose it. Of the \num{53} datasets in \textsc{WildlifeDatasets}/\textsc{MegaDescriptor}~\cite{cermak2024wildlifedatasets} only \textsc{AerialCattle2017} uses nadir views, because for cattle their coat pattern makes them separable. Red deer carry no such marking apart from the males' antlers, which makes nadir re-identification
harder.

\subsection{Selective prediction and frame quality} When evidence is weak it is often better to abstain than to force a label; selective classification formalizes this reject option~\cite{chow1970optimum,elyaniv2010foundations}, and modern variants pair a classifier with a confidence estimate that decides when to defer~\cite{geifman2017selective,hendrycks2017baseline}. These are general formulations rather than wildlife methods; the one treatment on aerial wildlife imagery we know of is the prior work on this same red deer data~\cite{softprior2026}. Here the analogous decision is \emph{which frames to classify at all}, since an occluded or badly posed crop carries no reliable demographic evidence, and forcing it into a class injects label noise~\cite{northcutt2021confident}. That work abstains per frame; we move the principle to the track level (\cref{sec:selective}).

\section{Method}
\label{sec:method}

\subsection{Data and evaluation protocol}
\label{sec:data}
We use the \texttt{matched} and \texttt{orthographic} subsets of the public BAMBI dataset~\cite{praschl2026bambi} (\url{https://github.com/bambi-eco/Dataset}): paired, undistorted RGB--thermal recordings from temperate Austrian forests and forest edges, flown nadir with DJI M3T/M30T drones at \SIrange{30}{60}{\meter} above ground. RTK multi-GNSS gives centimeter-level per-frame pose, so each frame can be used in the native camera perspective or re-projected onto the terrain as an orthophoto (``geo''), in which detections carry world coordinates. The dataset ships bounding-box \emph{tracks}, each with a species and, where identifiable, sex and age, plus an occlusion flag on the key frames. Red deer (\emph{Cervus elaphus}) is the richest demographic class: \num{1613} tracks and \num{26674} key frames, with sex annotated on \SI{41.0}{\percent} of tracks, age on \SI{67.6}{\percent} and an occlusion flag on \SI{49.5}{\percent} of key frames. Tracks and frames without a given label are unused by the corresponding head.

\paragraph{Three supervision sources.}
\emph{Species} labels come from a multi-species crop set (\num{653} roe deer, \num{6868} red deer, \num{4698} wild boar paired crops). \emph{Occlusion} labels are the per-key-frame visibility flags, giving \num{20531} labeled crops (\num{14744} clear, \num{5787} occluded). \emph{Sex} follows the strictest protocol: every matched crop was re-reviewed independently by three annotators free to abstain, and only the \num{3908} crops on which a majority agreed (\num{556} male, \num{3352} female/juvenile) were kept, so the sex head trains on a consensus-filtered set spanning the 2023--2024 antler cycle~\cite{softprior2026}, available at \url{https://zenodo.org/records/21061638}.

\paragraph{Evaluation.}
Detectors are split \emph{by flight}, every sixth flight held out, so no annotated \emph{track} is shared between training and validation. Individuals need not be disjoint, since the same animal can appear on more than one flight of an area, but the model never sees a track twice. Classification heads are evaluated leakage-free with GroupKFold \emph{by animal track}. The census pipeline is then evaluated on four further BAMBI flights, held out of every training set, from three test areas and denoted \emph{A1}, \emph{A2} (two flights over area~A, three years apart), \emph{B} and \emph{C}. These are the flights for which we hold field-verified demographic ground truth, so a per-individual census can be scored against what was present. Each census comes from one flight alone, and we do not link individuals across flights. Species is predicted for every detected animal; sex and life stage only for red deer, the one class with enough demographic annotation to support them.

\subsection{Detection and cross-modal tracking}
\label{sec:track}
\begin{figure*}[t]
\centering
\includegraphics[width=\textwidth]{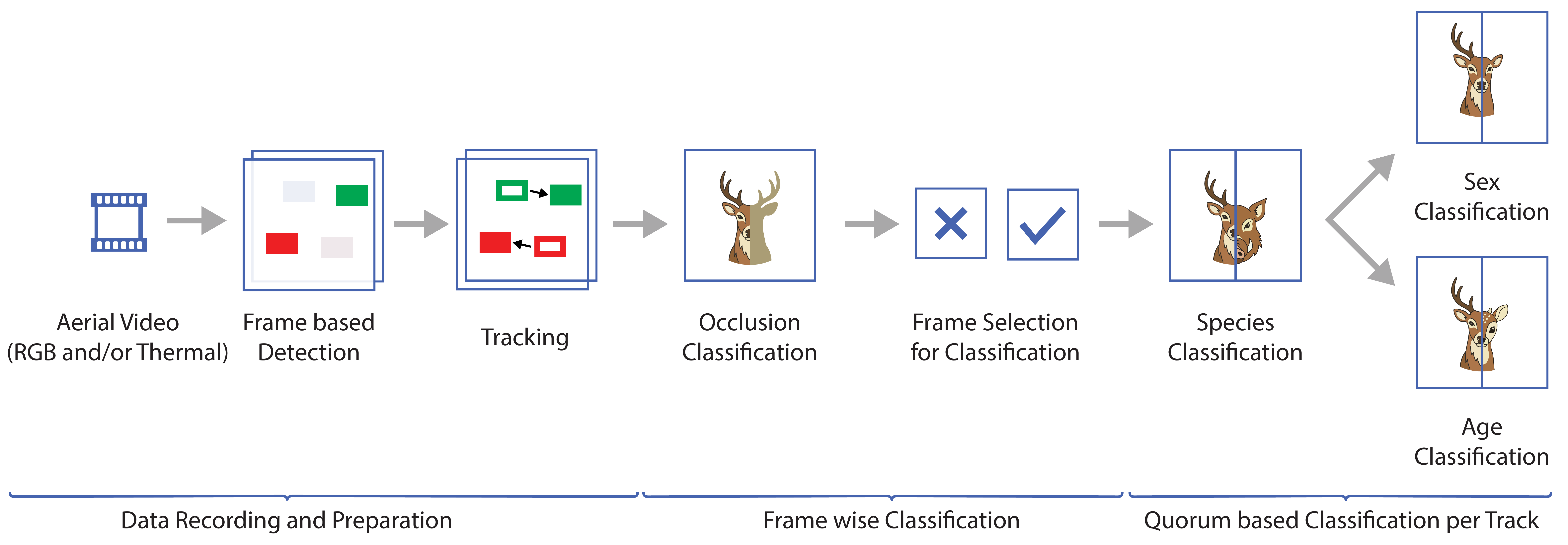}
\caption{Overview of the hierarchical, multimodal pipeline. Animals are detected and tracked in RGB and thermal video and cropped; each crop then passes occlusion filtering, species classification and demographic classification, with frame-level predictions aggregated per track by majority vote.}
\label{fig:pipeline}
\end{figure*}

\Cref{fig:pipeline} gives the overall flow: the synchronized RGB and thermal (TH) streams are detected, tracked and cropped, then passed through a \emph{hierarchical} classifier whose species and demographic stages run only on the frames an occlusion filter keeps. Rather than fuse the sensors at pixel level we fuse them in embedding space, so each supplies the cue it resolves best.

\paragraph{Detection.}
A single-class (\texttt{animal}) YOLO26x~\cite{yolo26} detector runs per modality at $1024\times1024$; for the orthographic tests the detectors are trained on content-cropped orthophotos instead (source $2048$\,px, typically around $1200$\,px after cropping) to avoid model-internal rescaling. At inference each panel is run at a deliberately low confidence, keeping every candidate in the pool $\mathcal{D}_m=\{b: s(b)\ge\tau_{\text{det}}\}$ with $\tau_{\text{det}}=0.01$ and $m\in\{\text{RGB},\text{TH}\}$, where $b$ is a candidate box and $s(b)$ its detector confidence. This pool is not what we report as detections; it is revisited later to fill gaps inside tracks that are already established.

\paragraph{Cross-modal association defines a track.}
Each modality is tracked independently with a constant-velocity SORT tracker~\cite{sort,kalman}, then registered at the \emph{track} level. From frames where both fire we least-squares fit a $2$D affine map
$T:\text{RGB}\!\to\!\text{TH}$,
\begin{equation}
  T(\mathbf{p}) = A\,\mathbf{p} + \mathbf{t}, \qquad
  (A,\mathbf{t}) = \arg\min_{A,\mathbf{t}} \sum_k \big\| A\,\mathbf{p}^{\text{RGB}}_k + \mathbf{t} - \mathbf{p}^{\text{TH}}_k \big\|_2^2 ,
  \label{eq:affine}
\end{equation}
where $\mathbf{p}^{\text{RGB}}_k$ and $\mathbf{p}^{\text{TH}}_k$ are the image centers of the $k$-th corresponding detection pair, $A$ the $2\times2$ linear part and $\mathbf{t}$ the translation. Each RGB track $i$ is compared to every thermal track $j$ by the \emph{median} inter-center distance over the frames they share,
\begin{equation}
  d(i,j) = \operatorname*{median}_{f \in \mathcal{F}_i \cap \mathcal{F}_j}
  \big\| T(\mathbf{c}^{\text{RGB}}_{i,f}) - \mathbf{c}^{\text{TH}}_{j,f} \big\|_2 ,
  \label{eq:match}
\end{equation}
with $\mathcal{F}_i$ the frames track $i$ covers and $\mathbf{c}_{i,f}$ its box center in frame $f$. A pair is accepted when the tracks share at least eight frames, when $d(i,j)$ falls below an empirically chosen gate of \num{28}\,px, and when both reach a mean detection confidence of \num{0.20}, which stops a match being built on a track that fires on shadow; assignment is one-to-one and greedy, smallest median distance first. This confirmation \emph{is} the track definition: a pair seen in both modalities is a real animal, a long and confident single-modality track is an animal one sensor cannot see, and short low-confidence tracks are sensor noise.

\paragraph{Thermal-anchored boxes.}
Once matched, the thermal box is taken as the reference size, being the looser of the two and so more reliably enclosing the whole animal, antlers included. The RGB box is redrawn at thermal size, anchored on the RGB centroid and refined by template matching as described by Praschl \etal~\cite{praschl2026bambi}. Tracks are established from confident detections only ($s(b)\!\ge\!0.2$); gaps are then filled from the pool $\mathcal{D}_m$ down to $s(b)\!\ge\!\tau_{\text{det}}$, snapping a low-confidence detection to the track when it lies near the position predicted from surrounding frames. Because a box is recovered only where a real detection supports it, none are hallucinated.

\subsection{Hierarchical demographic classification}
\label{sec:cls}\label{sec:selective}\label{sec:size}
Tracks yield frames of uneven quality, and forcing every frame into a label injects noise that
corrupts downstream classifiers~\cite{northcutt2021confident}. Prior work on this data abstains
on ambiguous crops~\cite{softprior2026}, but \emph{per frame}, discarding each detection in
isolation. We move the principle to the \emph{track}, selecting only the frames worth trusting
and aggregating across the survivors, so an animal is never discarded, only its unreliable
frames are.

\paragraph{From raw embeddings to classification.}
Every crop is embedded once with frozen DINOv3 ViT-H+ features~\cite{dinov3} in three \emph{input configurations}: \texttt{rgb} (\num{1280}-d), \texttt{thermal} (\num{1280}-d) and \texttt{combined} (\num{2560}-d, the concatenation). Each head (occlusion, species, sex) maps its input onto a \num{256}-d embedding with its own triplet-trained projection $\phi$, then classifies that embedding linearly. With anchor $a$, hardest positive $p$ and hardest negative $n$, $\phi$ is trained by batch-hard mining~\cite{triplet,hermans2017defense},
\begin{equation}
  \mathcal{L}_{\text{triplet}} = \big[\, \lVert \phi_a - \phi_p\rVert_2 - \lVert \phi_a - \phi_n\rVert_2 + \alpha \,\big]_{+},
  \label{eq:triplet}
\end{equation}
where $\phi_x$ is the projected embedding of crop $x$, $\alpha$ the margin and $[\,z\,]_{+}=\max(0,z)$. Crucially, \texttt{combined} concatenates the two feature vectors \emph{before} the projection, so the head learns a joint representation rather than combining two decisions. Reporting every stage for all three configurations isolates what each modality and their fusion contribute.

The labels are read in a fixed order: occlusion, then species, then sex. The occlusion head labels every crop \emph{clear} or \emph{occluded}, and only the clear frames $\mathcal{C}_i$ of individual $i$ continue. On $\mathcal{C}_i$ the species head votes, and its majority call fixes the species; this can stand as the final answer when the evidence is too weak for a demographic call. Sex reuses \emph{exactly} those frames. The sex head is binary, separating adult males from everything else, because the only cue visible from nadir is the antler: calves lose their natal spots within weeks, and thereafter a juvenile of either sex differs from an adult female only in size at this resolution. Its second class is therefore \female/\juv, split by size alone below. The head commits each clear crop $f$ to one class, giving a per-frame call $\hat{s}_f$, and the individual's sex is the majority call, $\hat{y}_i = \ind[\, \sum_{f\in\mathcal{C}_i} \ind[\hat{s}_f = \male] > \tfrac12 \lvert\mathcal{C}_i\rvert \,]$. Voting is what makes a noisy per-frame call safe to use: the antler is resolved only from some angles, so many frames of a true male look female, but a male that resolves his antlers on more than half of his clear frames still carries the vote. This is where we depart from the per-frame treatment of the same data in prior work~\cite{softprior2026}, which adds a confidence band around the decision boundary and lets the uncertain middle abstain~\cite{chow1970optimum,geifman2017selective}. We place the reject decision at the occlusion stage instead: filtering removes the frames on which the evidence is physically absent, and the vote absorbs the rest, so no individual is ever abstained on.

Sex leaves a juvenile inside the \female/\juv\ class, so life stage is recovered afterwards from body size. We switch to the orthorectified (``geo'') frames, whose box area is a metric body-size proxy, and restrict to the seasonal window in which adults and juveniles differ most. Because box tightness is not comparable between flights we compare only \emph{within} a flight, taking each individual's median track area $a_i$ and flagging low outliers with a robust median-absolute-deviation (MAD) $z$-score,
\begin{equation}
  z_i = 0.6745\,\frac{a_i - \operatorname{median}_j a_j}{\operatorname{MAD}_j a_j},
  \label{eq:mad}
\end{equation}
with the median and MAD taken over the individuals $j$ of that flight. Individual $i$ is called a juvenile only when \emph{both} conditions hold: it sits far below the flight median ($z_i<-2$), and its size gap to the next-smallest individual exceeds twice the interquartile range of the flight's areas. The second condition keeps the rule from firing on an adult that is merely the smallest of its herd~(\cref{sec:limits}).

\subsection{Appearance re-identification}
\label{sec:reid}
When an animal leaves the field of view and returns it is tracked afresh, so a census must
recognize it across tracks without double-counting. This is narrower than re-identification in its
usual sense: we match an animal to itself inside one flight, over minutes rather than seasons,
against at most a dozen candidates and a nearly unchanged background. It is duplicate suppression
for the census, not a general identity model, and the numbers below should be read that way.
Every detection is geo-referenced (\cref{sec:data}), so a new track at ground position
$\mathbf{x}_q$ is compared only against individuals last seen nearby, the candidate set
$\mathcal{N}_q=\{i:\lVert\mathbf{x}_q-\mathbf{x}_i\rVert_2\le r\}$, with $r$ the distance the
animal could plausibly have covered between sightings; an empty $\mathcal{N}_q$ means a new
animal at no appearance cost. Appearance is a per-track signature $\mathbf{s}_i$: clear crops are
embedded with DINOv3 ViT-H+~\cite{dinov3} (\num{1280}-d class token), $\ell_2$-normalized and
averaged within each modality, then concatenated into one \num{2560}-d vector compared by cosine.
The track goes to $\operatorname{arg\,max}_{i\in\mathcal{N}_q}\cos(\mathbf{s}_q,\mathbf{s}_i)$ when
that similarity clears a threshold $\theta$, and is new otherwise.

\FloatBarrier
\section{Results}
\label{sec:results}

\subsection{Detection}
\begin{table}[tb]
  \centering
  \caption{Single-class detection accuracy (flight-held-out validation). Thermal is the better
  detector in both domains, and orthorectification costs RGB \num{0.085} mAP@50 while leaving
  thermal essentially intact.}
  \label{tab:det}
  \setlength{\tabcolsep}{10pt}
  \begin{tabular}{llcc}
    \toprule
    Domain & Modality & mAP@50 & mAP@50--95 \\
    \midrule
    Perspective  & RGB     & 0.809 & 0.562 \\
    Perspective  & Thermal & 0.876 & \textbf{0.741} \\
    Orthographic & RGB     & 0.724 & 0.478 \\
    Orthographic & Thermal & \textbf{0.885} & 0.716 \\
    \bottomrule
  \end{tabular}
\end{table}

\begin{table}[tb]
  \centering
  \caption{What cross-modal confirmation contributes. Requiring the two sensors to agree cuts
  \num{94} raw tracks to \num{30}, every one a ground-truth individual, so confirmation is what
  keeps phantom animals out of the census. \emph{RGB} and \emph{TH} are the raw tracks each
  detector produces alone, \emph{Conf.}\ the pairs accepted by \cref{eq:match}, \emph{Unm.}\
  those without a partner, \emph{Adm.}\ those carried into the census as single-modality
  animals. On \emph{C} the ten are fragments of five animals that re-identification merges.}
  \label{tab:confirm}
  \setlength{\tabcolsep}{7pt}
  \begin{tabular}{lccccccc}
    \toprule
    Flight & Frames & RGB & TH & Conf. & Unm. & Adm. & Individuals \\
    \midrule
    A1 &  510 &  8 &  7 &  6 &  3 & 1 &  7 \\
    A2 & 1267 & 13 & 16 & 11 &  7 & 0 & 11 \\
    B  &  899 &  6 &  4 &  3 &  4 & 0 &  3 \\
    C  & 3126 & 16 & 24 & 10 & 20 & 0 &  5 \\
    \midrule
    Total & 5802 & 43 & 51 & \textbf{30} & 34 & 1 & \textbf{26} \\
    \bottomrule
  \end{tabular}
\end{table}

\label{sec:detres}
Thermal detects animals more reliably than RGB in both domains (\cref{tab:det}), raising mAP@50--95 in perspective imagery from 0.562 to 0.741. Orthorectification costs RGB a great deal, dropping mAP@50 from 0.809 to 0.724, most likely because the warp smears animals already small and low in contrast, while thermal is largely unaffected, its stricter mAP@50--95 falling only from 0.741 to 0.716. Thermal is thus robust across domains while RGB needs the native perspective, and that loss of detail propagates to sex classification later. Across the four test flights animals are found reliably, with three exceptions: on \emph{A2} the adult female D12 is missed for lack of contrast and D8 is not separated from her juvenile (\cref{fig:census}); on \emph{C} an adult female lying very close to the detected D3 is missed. Those three are absent from the census.

\paragraph{What cross-modal confirmation buys.}
Detection accuracy alone does not say how many of the resulting \emph{tracks} are animals, which is
what a census depends on. \Cref{tab:confirm} answers that on the four census flights, holding
\num{5802} synchronized RGB--thermal frame pairs and \num{26} detectable individuals, none in any
training set. The detectors produce \num{94} raw tracks between them, \num{43} in RGB and \num{51}
in thermal. Requiring the modalities to agree accepts \num{30}, every one of which belongs to a
ground-truth individual, registering tightly at median inter-center distances of
\numrange{0.8}{9.1}\,px inside the \num{28}\,px gate. Of the \num{34} tracks without a partner only
one is a real animal, the male A1~R8, whom thermal never picks up and whom the
length-and-confidence fallback correctly admits. That fallback is not reliable alone: applied
uniformly it would admit six more tracks that no ground-truth animal accounts for, five on
\emph{C}, where dense canopy yields long, confident single-modality tracks on background structure.

\subsection{Occlusion filtering}
\begin{table}[tb]
  \centering
  \caption{Per-frame occlusion classification on held-out tracks. Fusion makes the best filter,
  because a frame passes only when the animal is visible to both sensors: it removes
  \SI{71}{\percent} of the occluded frames (\emph{Occluded Rejected}), and \SI{86}{\percent} of what
  it keeps is genuinely clear (\emph{Retained Clear}).}
  \label{tab:occ}
  \setlength{\tabcolsep}{7pt}
  \begin{tabular}{lccc}
    \toprule
    Modality & Balanced Acc. (\%) & Occluded Rejected (\%) & Retained Clear (\%)  \\
    \midrule
    RGB      & 72 & 61 & 86 \\
    Thermal  & 70 & 63 & 83 \\
    Combined & \textbf{78} & \textbf{71} & \textbf{86} \\
    \bottomrule
  \end{tabular}
\end{table}

\label{sec:occres}
The occlusion head is the first gate of \cref{sec:selective}. On held-out tracks,
balanced accuracy is \SI{72}{\percent} for RGB, \SI{70}{\percent} for thermal and
\SI{78}{\percent} combined (\cref{tab:occ}). These are moderate numbers, because occlusion is a
hard and partly subjective call and \SI{72}{\percent} of frames are clear. We use the combined
head as a filter rather than a final label, and it discards
\SI{33}{\percent} of frames overall. Because the decision is per frame while the demographic call
is a per-track vote, rejecting this many frames is safe: every individual is still left with ample
clean, sexable observations.

\FloatBarrier
\subsection{Species classification}
\begin{table}[tb]
  \centering
  \caption{Per-frame \emph{species} accuracy (\%) with (\ding{51}) and without (\ding{55}) the
  upstream occlusion filter. Species is the one stage where fusion does not pay: thermal collapses
  on the small roe deer (\SI{7.7}{\percent}), and filtering makes RGB the best view overall
  (\SI{97.5}{\percent}), helping the roe deer most.}
  \label{tab:species}
  \setlength{\tabcolsep}{8pt}
  \begin{tabular}{llcccc}
    \toprule
    Occ.\ filter & Modality & Overall & Roe Deer & Red Deer & Wild Boar \\
    \midrule
    \multirow{3}{*}{\ding{51}}
      & RGB      & \textbf{97.5} & \textbf{92.3} & 97.4 & \textbf{98.5} \\
      & Thermal  & 81.9 &  7.7 & 81.6 & 90.8 \\
      & Combined & 90.4 & 69.2 & 90.1 & 94.7 \\
    \midrule
    \multirow{3}{*}{\ding{55}}
      & RGB      & 95.4 & 73.9 & 97.5 & 96.6 \\
      & Thermal  & 88.2 &  6.5 & 96.2 & 92.6 \\
      & Combined & 96.0 & 66.3 & \textbf{99.0} & 97.4 \\
    \bottomrule
  \end{tabular}
\end{table}

\label{sec:speciesres}
\Cref{tab:species} reports per-frame species accuracy for the three most common BAMBI ungulates
(roe deer, red deer, wild boar), with and without the upstream occlusion filter. Among the single
modalities RGB is clearly best (\SI{97.5}{\percent} overall with filtering), and the telling entry
is the roe-deer column under thermal: at \SI{7.7}{\percent} it essentially cannot pick out the small
roe deer, confusing it with the larger red deer once fine detail is lost to the warm blob. The
combined model (\SI{90.4}{\percent}) sits between the two, inheriting part of that weakness rather
than compensating for it, which suggests that on clean crops RGB already carries the
discriminative signal and thermal mainly adds noise at the species level.

The two blocks isolate the filter's effect. Without it the combined model leads, on
\SI{96.0}{\percent}; with it RGB leads, on \SI{97.5}{\percent}. The largest single change is roe
deer under RGB, from \SI{73.9}{\percent} to \SI{92.3}{\percent}, since occlusion costs a
low-contrast color crop more than a thermal blob. Filtering also removes frames, so the two blocks
are scored on different numbers of crops and are read for direction rather than exact magnitude.

\FloatBarrier
\subsection{Sex classification}
\begin{table}[tb]
  \centering
  \caption{Sex per individual: the majority-vote call of each head on the eight ground-truth adult
  males, grouped by which modality carries the cue. Either sensor alone sexes two of the eight, the
  fused head seven; on three of them both single modalities vote \female, so no late combination of
  their outputs could have recovered the male. The other \num{18} individuals (\num{16} adult
  females, two juveniles) are called correctly by every head and omitted for space.
  M${=}$\male, F${=}$\female/\juv; combined in \textbf{bold}.}
  \label{tab:sexmales}
  \setlength{\tabcolsep}{10pt}
  \begin{tabular}{llccc}
    \toprule
    Male Individual & Flight & RGB & Thermal & Combined \\
    \midrule
    \multicolumn{5}{l}{\itshape RGB carries the cue, thermal votes \female}\\
    D8 & A1 & M & F & \textbf{M} \\
    D3 & B  & M & F & \textbf{M} \\
    \midrule
    \multicolumn{5}{l}{\itshape thermal carries the cue, RGB votes \female}\\
    D4  & A2 & F & M & \textbf{M} \\
    D11 & A2 & F & M & \textbf{M} \\
    \midrule
    \multicolumn{5}{l}{\itshape neither single modality carries it; only fusion recovers}\\
    D1  & A1 & F & F & \textbf{M} \\
    D10 & A2 & F & F & \textbf{M} \\
    D5  & C  & F & F & \textbf{M} \\
    \midrule
    \multicolumn{5}{l}{\itshape the one miss}\\
    D4 & A1 & F & F & \textbf{F} \\
    \bottomrule
  \end{tabular}
\end{table}

\begin{figure}[tb]
  \centering
  \includegraphics[width=\linewidth]{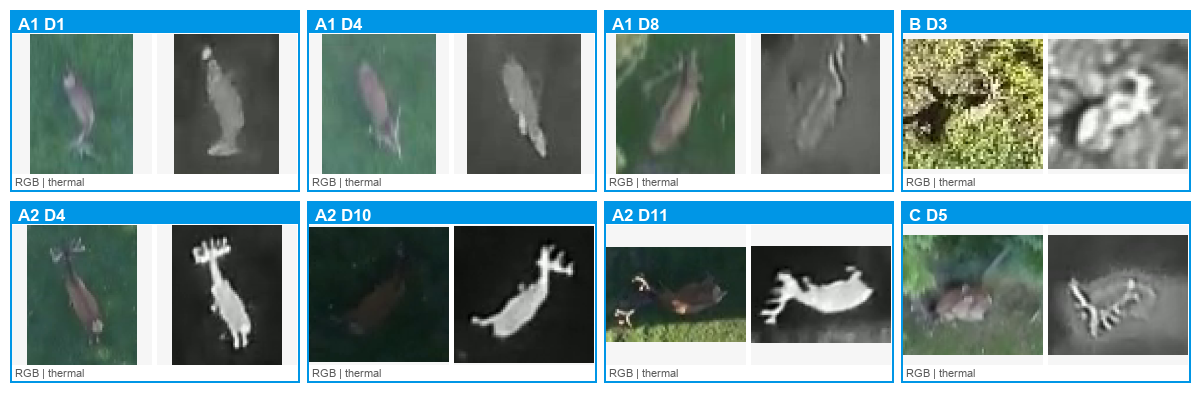}
  \caption{The eight ground-truth males, each at its clearest frame as an RGB (left) and thermal
  (right) crop. Thermal resolves the antlers for the velvet bucks (\eg\ A2~D4, D11) but not the
  hardened-antler males; the combined head recovers all but A1~D4.}
  \label{fig:malesamples}
\end{figure}

\begin{figure}[tb]
  \centering
  \includegraphics[width=\linewidth]{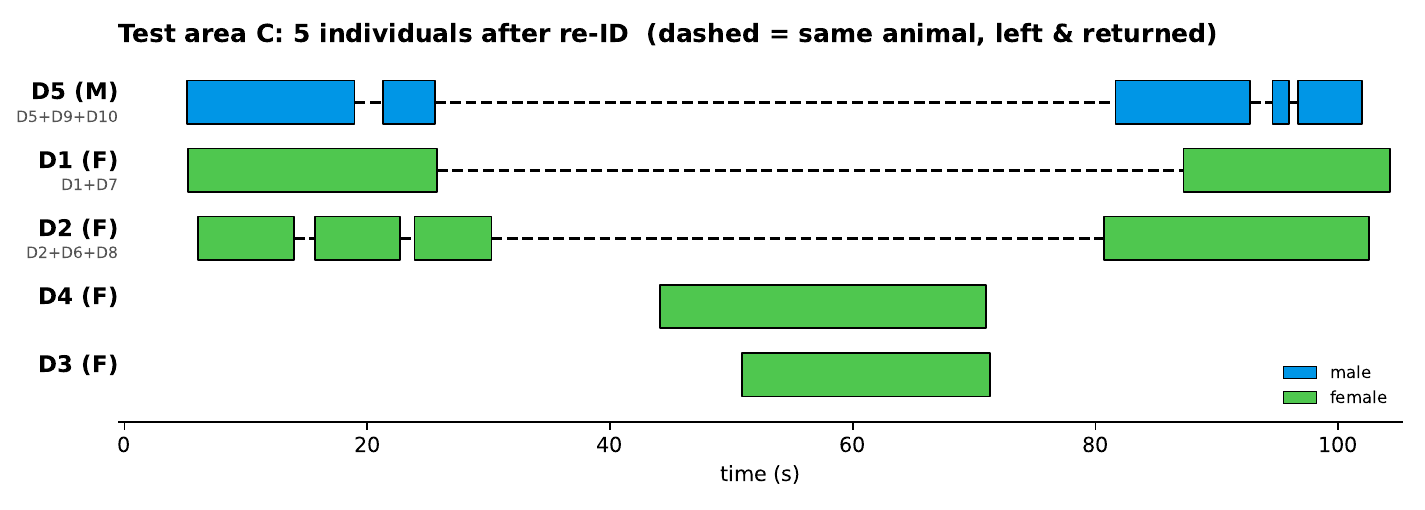}
  \caption{Re-identification on \emph{C}: thirteen raw tracks collapse to the five ground-truth
  individuals. Dashed segments mark an animal leaving and returning; the male D5 is
  recovered over four re-entries by pooling RGB (early) and thermal (late) antlers.}
  \label{fig:reid}
\end{figure}

\begin{figure}[tb]
  \centering
  \includegraphics[width=\linewidth]{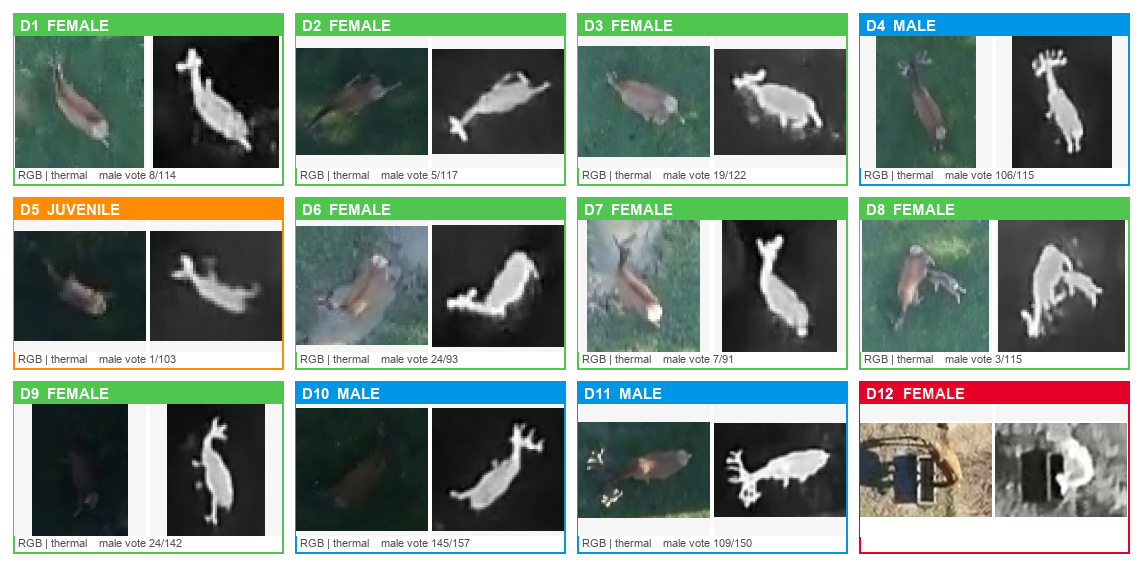}
  \caption{The \emph{A2} census: each re-identified individual with its RGB and thermal crop, the
  fused call and its male-vote fraction over clear frames. All eleven match the field ground truth:
  males D4 (\num{106}/\num{115}), D10 (\num{145}/\num{157}), D11 (\num{109}/\num{150}), and the
  size-flagged \juv\ D5. D10 is the fusion-recovery case, both single modalities voting \female\
  (RGB \num{17}/\num{157}, thermal \num{59}/\num{157}). No \female\ is over-called, not even D6
  (\num{24}/\num{93}). D12 is never detected.}
  \label{fig:census}
\end{figure}

\label{sec:sexres}

On top of species we classify sex (\male\ vs \female/\juv) on red deer. The three DINOv3 heads are
trained \emph{once}, on the three-annotator consensus set (\cref{sec:data}), then evaluated frozen
on the four held-out flights, each animal aggregated by majority vote over its non-occluded
frames: one classifier and one rule, scored once per flight.

The result is decided per individual. The combined head calls all \num{18} \female/\juv\
individuals correctly, that is \num{16} adult females and both juveniles, plus seven of the eight
adult males, missing only A1~D4; either sensor alone recovers just two of the eight. Fusion wins
because the single-modality failures are \emph{complementary} (\cref{tab:sexmales}): on two males
the antler cue survives only in RGB, on two others only in thermal, and on three males \emph{both}
vote \female\ yet the combined head still recovers \male\ (\cref{fig:census}). Those last three are the informative
cases: recovering them needs the joint representation itself, so fusion buys more here than any
late combination of the two outputs could.
\Cref{fig:malesamples} shows the clearest crop of each male.

Which sensor carries a male follows the antler cycle, and so the acquisition date: velvet antlers
are vascularized and thermally hot and light up the thermal view, but once the velvet is shed the
hardened antler goes thermally dead and only RGB resolves
it~\cite{potrapeluk2021thermography,lincoln1992antlers}. The flights span the cycle, from summer
velvet (B 2023-07-27, C 2026-06-25, A2 2026-06-26) to autumn hardened antler (A1 2023-09-20), so
the modality carrying sex is intermittent while the fused view stays reliable. On \emph{C}, canopy
cover and a later fly-back over the same ground split the five individuals into thirteen raw
tracks, eleven on D5, D2 and D1. Re-identification stitches these back across eight re-entries
(\cref{fig:reid}); the returners have moved, so appearance rather than position matches them. That
keeps the single male from being counted as several animals and pools his scattered frames, so the
intermittent cue still suffices to sex him.

\FloatBarrier
\subsection{Size and the juvenile}
\begin{figure}[tb]
  \centering
  \includegraphics[width=\linewidth]{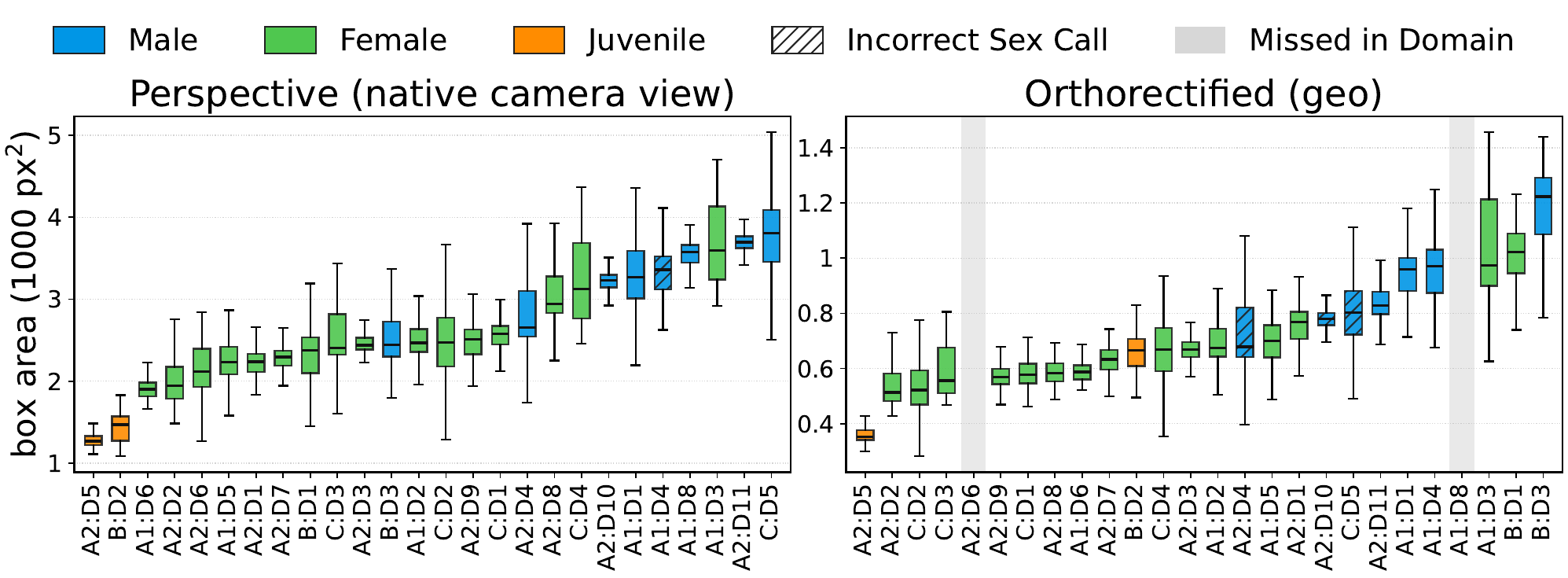}
  \caption{Per-individual thermal box areas, perspective (left) and orthorectified (right), sorted
  by median size, filled by field ground truth and sharing identity labels across domains. \male\
  and \female\ overlap over the whole range, so box area is a \emph{life-stage} cue, not a sex cue.
  Hatching marks an incorrect sex call, one in perspective and three after orthorectification, each a
  \male\ read as \female. Grey slots are individuals missed in that domain (A1~D8 from its RGB box).}
  \label{fig:size}
\end{figure}

\begin{figure}[tb]
  \centering
  \includegraphics[width=0.85\linewidth]{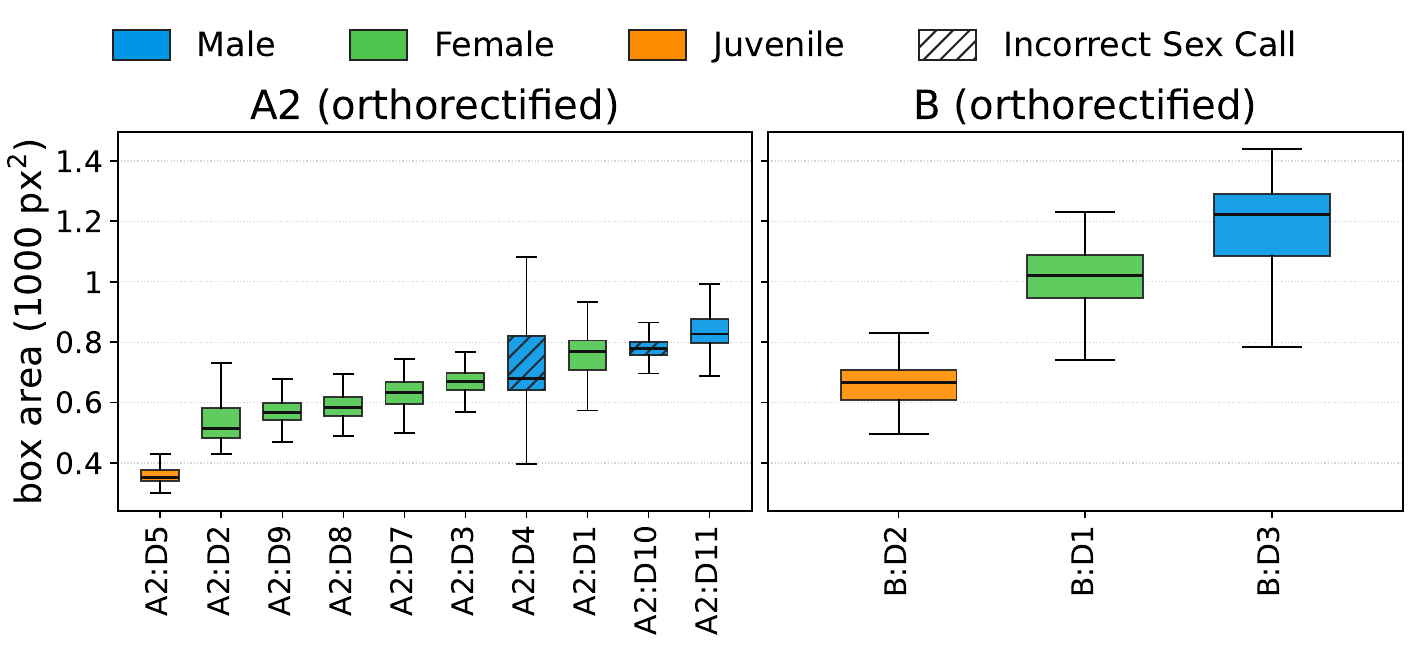}
  \caption{Orthographic thermal box areas for the two flights containing a juvenile, \emph{A2} and
  \emph{B}, on a shared axis, filled by field ground truth; hatching marks an incorrect sex call.
  Within each flight the juvenile (orange) is the clear low outlier ($0.53\times$ and $0.60\times$
  the flight-cohort median), yet the \emph{B} juvenile's box ($\sim\!673$\,px$^2$) still lands in
  \emph{A2}'s \female\ range, its detections carrying looser padding.}
  \label{fig:sizeperflight}
\end{figure}

Two of the four flights contain a juvenile, and the box-area outlier test flags each cleanly. The
smallest animal is D5 on \emph{A2} and D2 on \emph{B}; each sits far below its own flight cohort,
at $0.32\times$ and $0.60\times$ the cohort median in the native view, with Cohen's
$d>3$~\cite{cohen1988statistical} and no adult close to it in size, and field ground truth
confirms both as juveniles. \emph{A1} and \emph{C} hold no such outlier, their smallest animals
lying on a smooth size continuum, so the test does not over-fire.

Orthographic size is also \emph{metric}, so it should in principle compare across flights and
altitudes. That does not yet hold: on \emph{B} the thermal signal sits on a noisier background and
the boxes are visibly looser, so the \emph{B} juvenile's box ($\sim\!673$\,px$^2$) drifts up into
the \female\ range of \emph{A2} even though it is the smallest animal in its own flight
(\cref{fig:size,fig:sizeperflight}). Loose-versus-tight box variation carries straight into the
metric size (\cref{sec:limits}), so we read life stage \emph{per flight} rather than from one
global threshold, and leave the cross-flight metric as work in progress.

\FloatBarrier
\section{Discussion and Conclusion}
\label{sec:discussion}\label{sec:conclusion}\label{sec:limits}
We turn dual-stream UAV video of red deer herds into an individual-level demographic census by
fusing RGB and thermal at every stage, and which modality to trust proves situational. Thermal is
the better detector and the only view surviving cast shadow, yet it confuses the small roe deer
with red deer (\SI{7.7}{\percent} species accuracy) and its sex accuracy swings from
\SI{37}{\percent} to \SI{91}{\percent} with the
environment; RGB carries the antler cue but loses shaded animals. Alone, each recovers only two of
the eight males the fused head resolves to seven. Selecting frames rather than rejecting animals
keeps that call clean; a box-size biometric supplies the life stage appearance cannot.

\paragraph{Beyond axis-aligned boxes.}
The geometric cue, box area as a life-stage proxy, rests on an axis-aligned box. A deer seen from above is elongated and lies at an arbitrary heading, so a diagonally oriented animal forces the box to include more background than an axis-parallel one, and the area varies with orientation as much as with body size. That weakens the juvenile-to-adult separation and feeds heading-dependent background into crops. Oriented boxes would give tighter, pose-aligned footprints~\cite{xia2018dota,ding2019learning}, and a head--tail keypoint pair would estimate orientation and length directly~\cite{mathis2018deeplabcut,ng2022animal}; orthorectified, that distance is a metric size cue that also flags curled or foreshortened frames. That is the clearest next step.

\paragraph{Remaining limitations.}
Three limitations remain. DINOv3 is not discriminative enough for individual aerial deer without mean-centering, and only the strongest re-identification pairs survive. Males are rare enough to leave single-modality heads prone to minority errors. And two animals sit near the boundary: the male A2~D10, whom only fusion recovers, and the female C~D1 lying down, whose splayed posture reads as faintly male in the joint embedding (\SI{46}{\percent}) although both single modalities call her correctly.

\paragraph{What this changes for aerial surveys.}
A drone flight today may return a count, and the demographic reading that management acts on is added afterwards by hand, on a subsample of flights or not at all. Automating it lets composition be read on every flight, which is the frequency at which sex ratios and age structure stop being an occasional estimate and become trend data, and every call stays auditable in the frames and the vote behind it. What a survey reports changes with it: not one population number but a list of individuals, each with a species, a sex, a life stage and a confidence that a manager can question and carry into a population model. Red deer are an unusually legible case, so what transfers is the design rather than these accuracies: a demographic label is better taken as a decision over a track and two sensors than over one crop, and the evidence for it is seasonal, so a pipeline committed to one sensor will fail at some point in the year.

\smallskip\noindent \textbf{Availability.} Models: \href{https://huggingface.co/collections/cpraschl/models-when-one-modality-is-not-enough}{cpraschl/models-when-one-modality-is-not-enough}; data: \href{https://doi.org/10.5281/zenodo.21299620}{10.5281/zenodo.21299620}.

\bibliographystyle{splncs04}
\bibliography{main}

\end{document}